# Classification of Non-analyzable Word Types in Web Documents to Implement an Effective Korean E-Learning System

**Sang-Taek Park*, Ae-Lim Ahn*, Eric Laporte**, Jee-Sun Nam***
*DICORA, Hankuk University of Foreign Studies, Korea/ **LIGM, Université Paris-Est, France
fournine@gmail.com, aelimahn@gmail.com, Eric.Laporte@univ-paris-est.fr, namjs@hufs.ac.kr

## 1. Introduction

With the significant growth of IT infrastructure around the world, Internet is causing changes in various fields of Education. Some people need specific training to get high-quality language skills, or have limited exposure to foreign languages, and Internet can meet their needs conveniently. Starting with English, research on developing e-learning systems has gradually increased since late 1990s and developed effective usage of Internet in language education. Unlike English, e-learning for Korean was left behind because most interest in language learning in Korea was concentrated on English. However, since the early 2000s, the reputation of Korea has grown and significantly more foreigners want to learn Korean. As evidence of this, TOPIK, a Korean certification test for non-native speakers, has 14% more takers than a few years ago, and its rate increases each year. Therefore, research on Korean e-learning should be conducted through various fields such as linguistics, Korean education and computational science.

"E-learning is just-in-time education integrated with high velocity value chains. It is delivery of individualized, comprehensive, dynamic learning content in real time, aiding the development of communities of knowledge, linking learners and practitioners with experts" (DGWP, 2000). E-learning systems should deliver contents which reflect various phenomena of the language as it is used. That is, the more various contents the system can offer, the more developed it is. In this view, current Korean e-learning systems are at a quite basic level. There are four major e-learning sites for Korean:

국립국어원 바른소리 (http://www.korean.go.kr/hangeul/cpron/main.htm),

문화관광부 한글 (http://www.mct.go.kr/Hangeul),

한국외국인근로자센터 생활한국어 (http://www.mct.go.kr/Hangeul),

KBS World Radio (http://rki.kbs.co.kr/learn_korean/lessons/i_index.htm).

They provide useful expressions with text, sound, image, video etc. They focus on contents for people learning academic Korean and on basic travel phrases. Sites only differ in the way they provide the contents, but the contents are very limited. For high-level learners, this type of offer is not satisfying. In addition to formal Korean, e-learning systems that would include real-world Korean expressions such as those in web documents, mobile text messages, or twitter posts, would be useful for such learners.

Because of the fast development of personal blogs, twitter posts and individual homepages, much information is provided through them. This information is useful for foreigners as well as Korean speakers, and it is necessary for them to communicate with Korean people through this part of the web. The linguistic characteristics of these informal documents are quite different from those of formal documents such as academic text books or online newspaper articles. That is why many Korean learners have trouble in understanding these documents. Consider the following example:

(1a) 영화 **잼**있어요 [*Yeonghwa **jaem** iss-eo-yo*]. (= The movie is interesting)

(1b) **테레비**가 너무 작아요 [***telebi**-ga nemu jag-a-yo*]. (= The television is too small)

These two sentences include non-standard words. In (1a), '잼[*jeam*]' is a shortened form of '재미[*jeami*],' which means 'interest'. This is not a standard word, but recently it is used very frequently among young people. In (1b), an English loan word '테레비[*televi*]', which corresponds to 'television', is used, whereas its standard transliteration into Korean is '텔레비전[*tellebijeon*]'. It can also be transliterated into '티비[*tibi*]', '테레비죤[*telebijon*]' or '텔레비[*tellebi*]' etc.

Therefore, if Korean learners do not know this variation, it is not easy for them to understand Korean text including these expressions. These problems occur more and more frequently, we need to develop a systematic device to treat them in an e-learning system.

In this paper, we construct two types of corpora: one is made of formal documents like online news articles; the other is made of informal documents like customer reviews about new products in web blogs. By comparing these corpora, we show how expressions differ in these two types of corpora. We survey the main characteristics of the informal corpus. Given that a significant proportion of text is informal, we propose *Local Grammar Graphs* (LGG)[1] as an appropriate model to treat them effectively in Korean e-learning systems.

## 2. Comparative data between standard Korean and online Korean

In this chapter, we attempt to analyze non-standard word types frequently observed in web documents. We constructed two types of corpora: from formal Korean documents such as newspaper articles, and from informal Korean documents such as online review texts or personal blogs. We compare them to show how they differ in word formation and usage. Among web documents, we contrast formal Korean texts, like news articles or objective academic texts that require an accurate language style, with informal documents, like customer reviews or personal blogs where users can bring up their subjective views. News articles have to include standard expressions, while customer reviews or personal blogs include many colloquial and emotional expressions. For several reasons, non-standard word errors may appear in the process, which may in turn disturb learners of Korean as a second language. Without making sense of non-standard words, non-native Korean learners cannot communicate with other Korean speakers.

First of all, we observe the result obtained by applying Korean lexical analyzer *Geuljabi*[2] to these two types of corpora. Based on this program, non-standard words cannot be analyzed, since they are absent from the Machine-Reable Dictionary the program uses for morphological analysis. Let us compare the number of these 'non-analyzable words' from our two types of corpora:

| | **Corpus A** | **Corpus B** |
|---|---|---|
| **Text Type** | News articles (Social issue Section) | Customer reviews (About Cosmetic Product) |
| **Corpus Size** | 10,488 tokens | 10,608 tokens |
| **Types** | 3,967 | 3,792 |
| **non-analyzable words** | 152 types (about 3.8%) | 1,062 types (about 27%) |

**Table 1. Comparative Data of the two types of corpora**

[1] The model of Local Grammar Graphs (LGG) is proposed by Maurice Gross (1997) to formalize natural language with Finite-State-Automata-based Grammars for automated Natural Language Processing systems. It can effectively formalize linguistic expressions that violate general syntactic rules and are difficult to describe in the usual ways. The LGG model allows us to describe partial phenomena rather than whole sentences and to analyze and process compound nouns, frozen idioms, collocations, and synonymic relations between certain phrases.

[2] For more detailed information, see "www.sejong.or.kr"

Corpus A, which is made of news articles, consists of a total of 3,967 types, among which only 152 types are unanalyzable. It means only 3.8% of all word types are not recognized and more than 96% are analyzed. And even these 3.8% include typing errors or wrong space problems which are obviously unintentional. Besides, many proper nouns are not recognized because they are missing from the electronic dictionary. Therefore, it seems that these non-analyzable words are accidental or result from a lack of information in the dictionary. On the contrary, Corpus B, which is made of customer reviews, consists of a total of 3,782 types, among which 1,062 types are unanalyzable, accounting for 27% of all types. It means almost one third of types is not recognized. These non-standard words belong basically to two types: some are caused by a lack of knowledge of online writers in Korean spelling or grammar. Others come up when online writers intentionally break Korean spelling or writing rules. These intentionally deformed words sometimes bring a new effect in meaning or sentiment interpretation. By force of being used frequently among young people, some of them become new lexical items which should be registered in the Korean dictionary. Therefore, we cannot ignore them, especially when they become a big obstacle for foreigners who read web documents such as personal blogs or customer reviews.

## 3. Typology of non-standard expressions

In informal documents there are many types of non-standard words. We here classify the expressions which writers intentionally create by breaking Korean spelling or writing rules.

### [1] Intentionally missing spaces between words

In Korean like in English, spaces between words are necessary, except, for example, between a noun and a case-marking postposition (equivalent to a preposition in English). However, people often omit it in online texts out of laziness and convenience:

(2a) in online texts: 색깔이예뻐요 [*saegkkal-i-yeppeo-yo*] (=The color is good.)
(2b) in formal texts: 색깔이 예뻐요 [*saegkkal-i* **(space)** *yeppeo-yo*] (=The color is good.)

In many cases, it does not make confusion for native speakers but foreign learners can have problem in reading and understanding them. Optional spacing, such as in the complex verb "뛰어가다/뛰어 가다 (*ttuio-gada* = run and go)", is a different situation.

### [2] Abbreviations

People use abbreviated forms in online textsin order to save their effort and time:

(3a) [Shortened form] 이 상품을 **강추**합니다 [*I sangpum-eul* ***gangchu**-habnida*].
(=I strongly recommend this.)
(3b) [Standard form] = 이 상품을 **강력 추천**합니다 [*I sangpum-eul* ***ganglyeog chucheon**-habnida*].

(4a) [Shortened form] 효과가 **넘** 좋아요 [*Hyogwa-ga* ***neom*** *joh-a-yo*]. (=It has a very good effect.)
(4b) [Standard form] = 효과가 **너무** 좋아요 [*Hyogwa-ga* ***neomu*** *joh-a-yo*].

In example (3), '강추[*gangchu*]' is a shortened form of '강력 추천[*ganglyeog chucheon*]', which means 'strong recommendation'. In this case, the abbreviated word is formed by the initial syllables of two words. In (4), the abbreviation changes a single, two-syllable word '너무[*neomu*]', which means 'very', into a one-syllable word '넘[*neom*]'.

## [3] Expressions with deviant spelling

Especially young people do not care of spelling, sometimes on purpose, for fun, or out of feeling free. As the phenomenon of spelling deviance becomes more frequent, it will give big difficulty to foreign learners:

(5a) [Wrongly spelled form] 안녕하**세욤** [*Annyeongha**seyom***]. (=Hi)
(5b) [Standard form] = 안녕하**세요** [*Annyeongha**seyo***].

The Korean word corresponding to 'Hi' in English is 안녕하세요[*Annyeonghase-yo*] like in (5b). In Korean, '-세요[-se*yo*]' is a common predicative suffix, but many young people purposely change its spelling to '-세욤[-*seyom*]', '-세여[-*seyeo*]', '-셈[-*sem*]' etc.

## [4] Diversity in the spelling of loan words

There are many loan words from English in modern Korean. There exist an official regulation of transliteration of loan words, but it is not well respected by Korean speakers. It makes foreign learners confused, even sometimes Korean speakers as well. People tend to write a loan word as they pronounce it without considering the norms, and therefore sometimes for one English word we can observe many Korean transliterations. For instance:

(6a) [one transliteration] **쵸콜렛**향기 [***Chyokolles*** *hyanggi*] (= **Chocolate** perfume)
(6b) [Official regulation] **= 초콜릿**향기 [***Chokollis*** *hyanggi*]

Because there is no genuine Korean word to substitute 'chocolate', some transliteration must be used. This word is highly frequent in Korean; however, it can be transliterated in many ways such as '쪼꼬렛 [*jjokkoles*]', '초코렛 [*chokoles*]' and '초코레트 [*chokoleteu*]' etc. Some forms are used more frequently than the standard '초콜릿 [*chokollis*]'.

In principle, loan words are supposed to be transcribed into Korean in a way that renders faithfully the original words and fits the Korean phonological structure. To be specific, [f] and [p] should be transliterated by the Korean consonant 'ㅍ', and a complex consonant cannot appear in the first sound of loan words. In particular, a fortis, unvoiced consonant in English is not transcribed into a double consonant in Korean. This rule does not prevent semantic differentiation.

## [5] Newly coined words

Recently most newly coined words have come from online texts. In the cyberspace, communicating with one's friends informally using new words makes people feel closer. However, some of these words spread fast out of personal cyberspaces. They become common in the daily communication of the young generation. Therefore, foreign learners may want to learn them to communicate with them without difficulty. Consider:

(7a) [Newly coined form] **짱** 멋있다 [***Jjang*** *meos-issda*]. (=It's **very** nice.)
(7b) [Normal form] **= 진짜** 멋있다 [***Jinjja*** *meos-issda*].

To express emphasis, people usually insert an adverb. The new word '짱[*Jjang*]' means 'very', and can be replaced by '진짜[*Jinjja*]'. Besides, people have a tendency to use transliterated English words which were not usually used in Korean before, such as:

(8a) [New term with English word] 모양이 **심플하다** [*mo-yang-i* ***simpeul-hada***](=The shape is **simple**)
(8b) [Genuine Korean term] = 모양이 **단순하다** [*mo-yang-i* ***dansun-hada***].

The new term '심플하다[*simpeulhada*]' is made of the English word 'simple' and the Korean adjectival suffix '-하다[*-hada*]'.

### [6] Emoticons

Emoticons are usually used to make emotional phrases more expressive or to replace them. Since they have different forms according to the language and culture, foreigners may wish to learn the emoticons in use among Koreans:

(9a) Fun or joy: **ㅋㅋ**, **ㅎㅎ**, etc
(9b) Sadness: **ㅠㅠ**, **ㅜ_ㅜ**, etc
(9c) Agreement: **ㅇㅇ**, **ㅇㅋ**, etc
(9d) Others: *-*, @@, gg, oo, etc. [what do they mean?]

## 4. Applying LGGs by using the Unitex system

We here suggest a proper tool to deal with the problems of non-standard words effectively: the Local Grammar Graph (LGG) formalism. The LGG model, proposed by Gross (1997, 1999), is appropriate for effective formalization of linguistic forms that violate general syntactic rules and are difficult to describe in the usual ways. This formalism provides an effective mechanism for treating these "local" expressions and allowed us to implementa process which automatically transforms non-standard words into standard ones, and which could be a part of an e-learning system for foreign learners.

We built LGGs recognizing the non-standard words we presented in this paper and restoring the standard forms in the case of [1] Abbreviations, [2] Deviant spelling, [3] Alternative transliterations of loan words, [4] Newly coined words, [5] Emoticons. These LGGs could be constructed by using the Unitex Graph Editor and compiled into finite-state transducers (FSTs) (Paumier 2003: http://www-igm.univ-mlv.fr/~unitex/). The following screenshots show (1) an example of a graph, (2) a sample of the English Dictionary of simple, compound and unknown words included in Unitex, (3) a Concordance of an English text produced by Unitex:

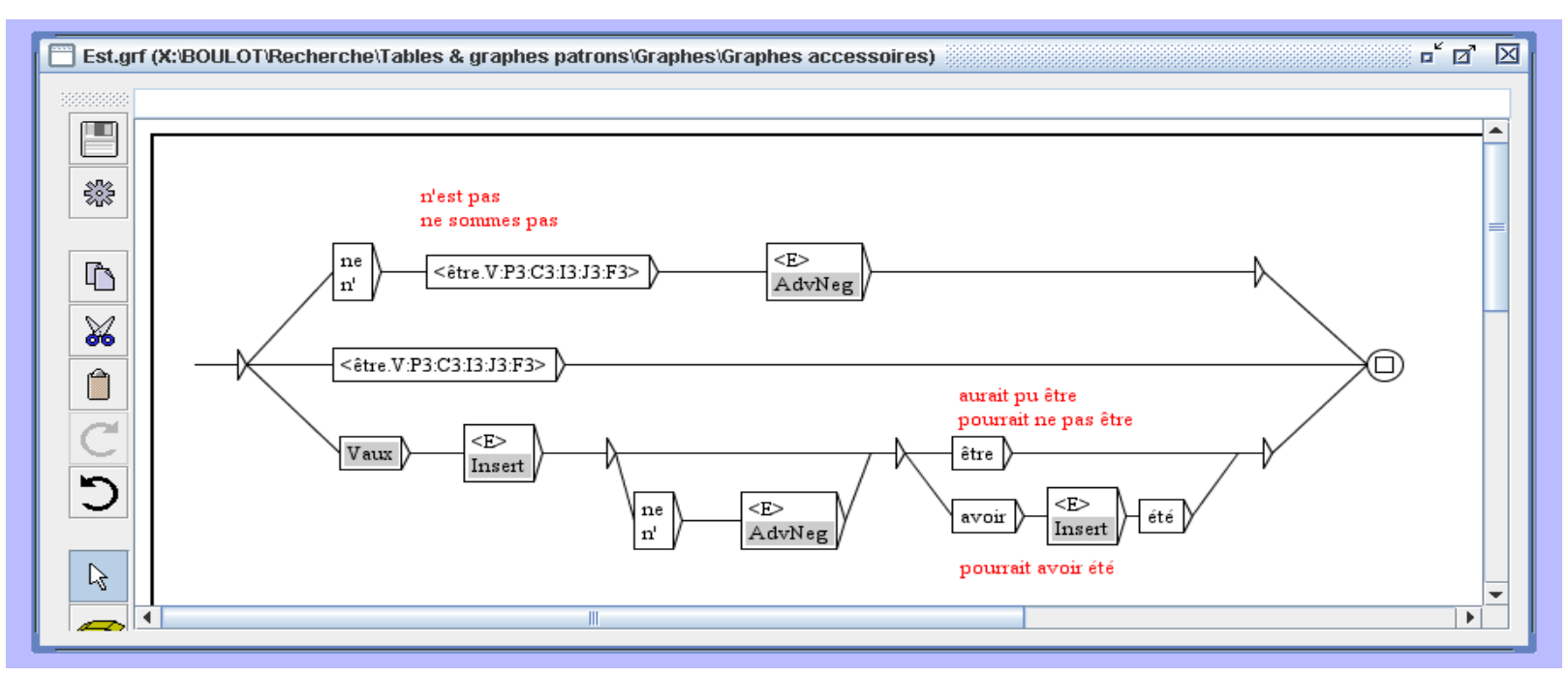

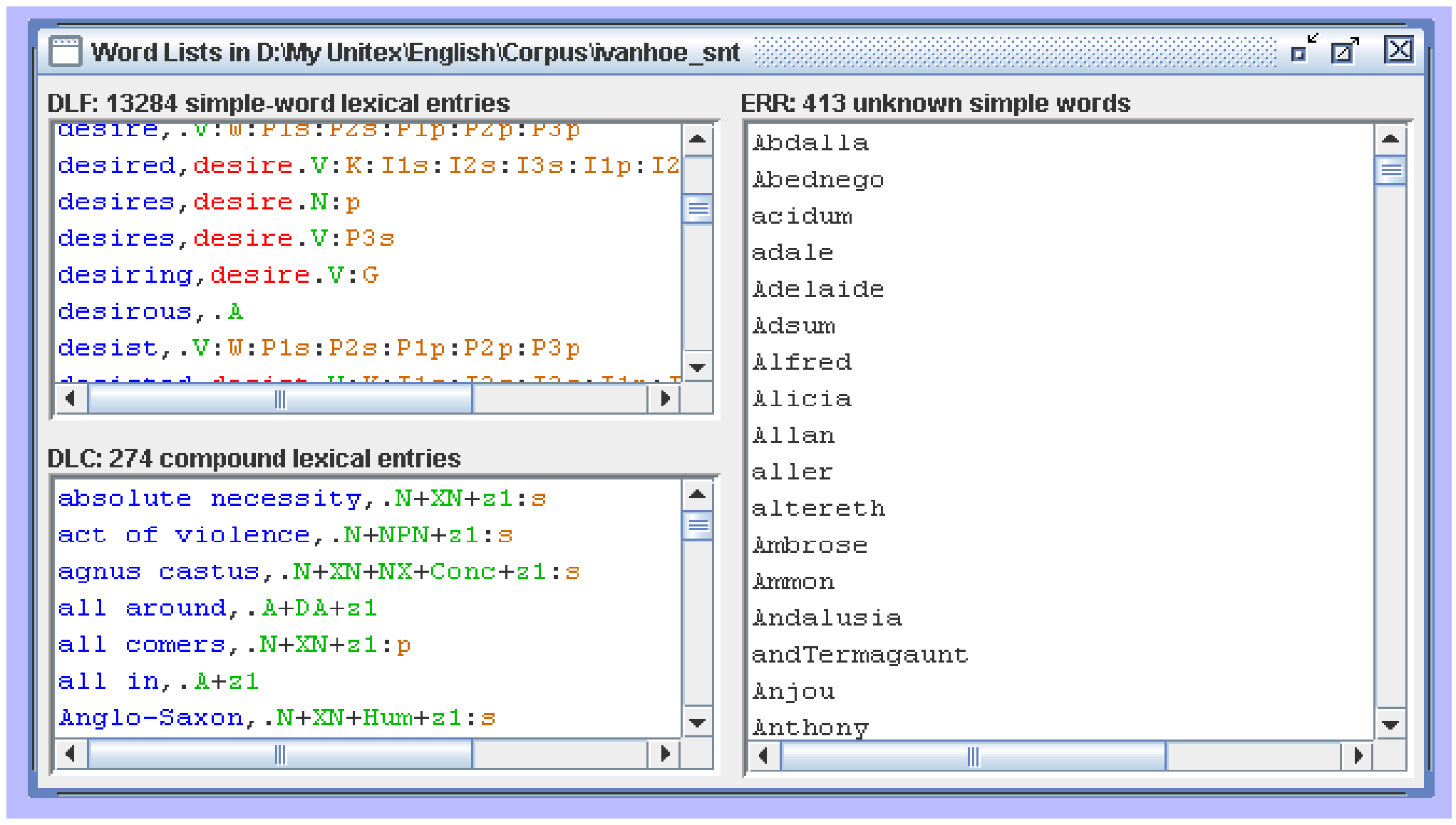

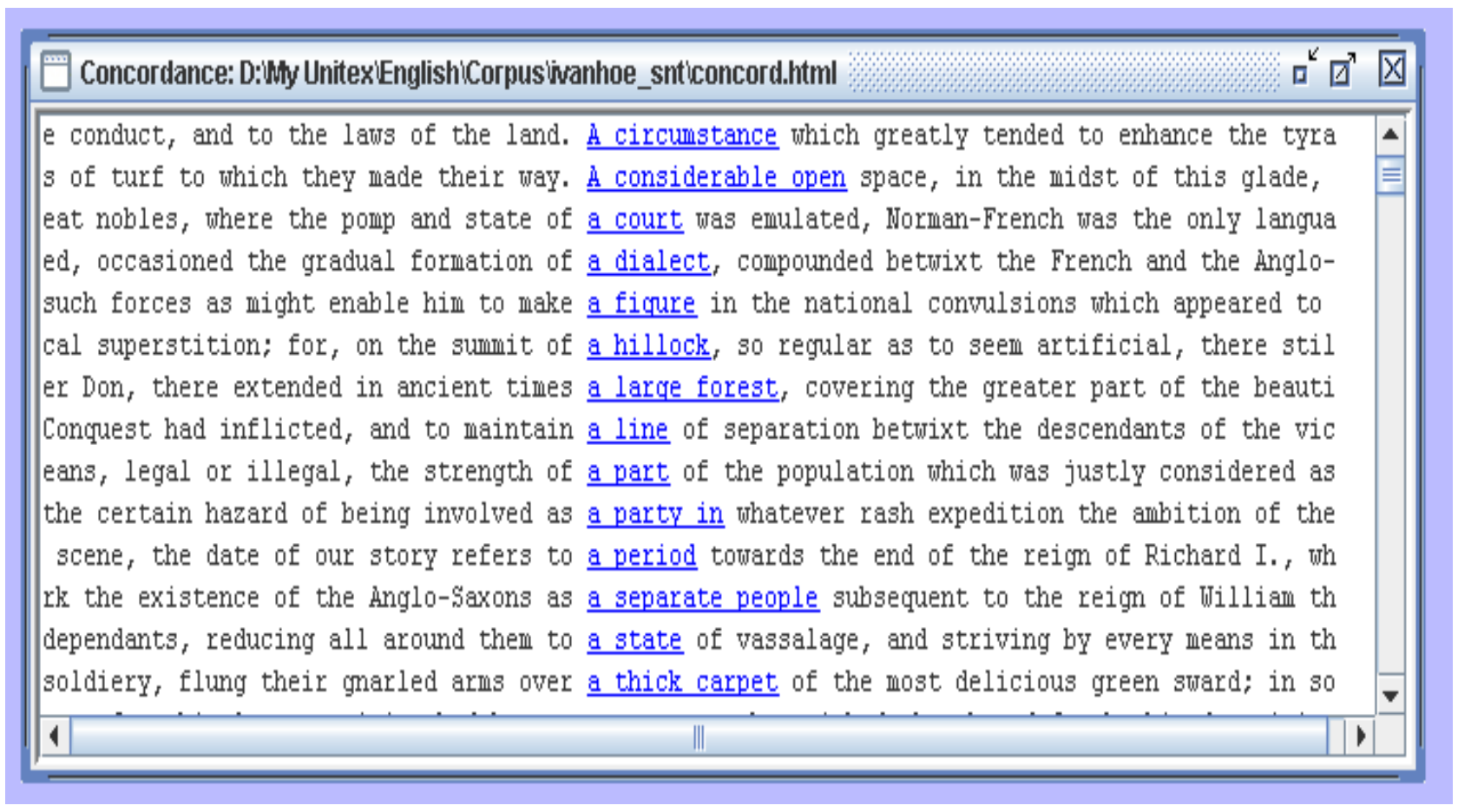

As the Unitex program is compatible with several languages such as English, French, German, Korean, Thai, Italian, Greek or Russian, the LGGs we construct for each language can be used in monolingual or bilingual e-learning systems in an effective way:

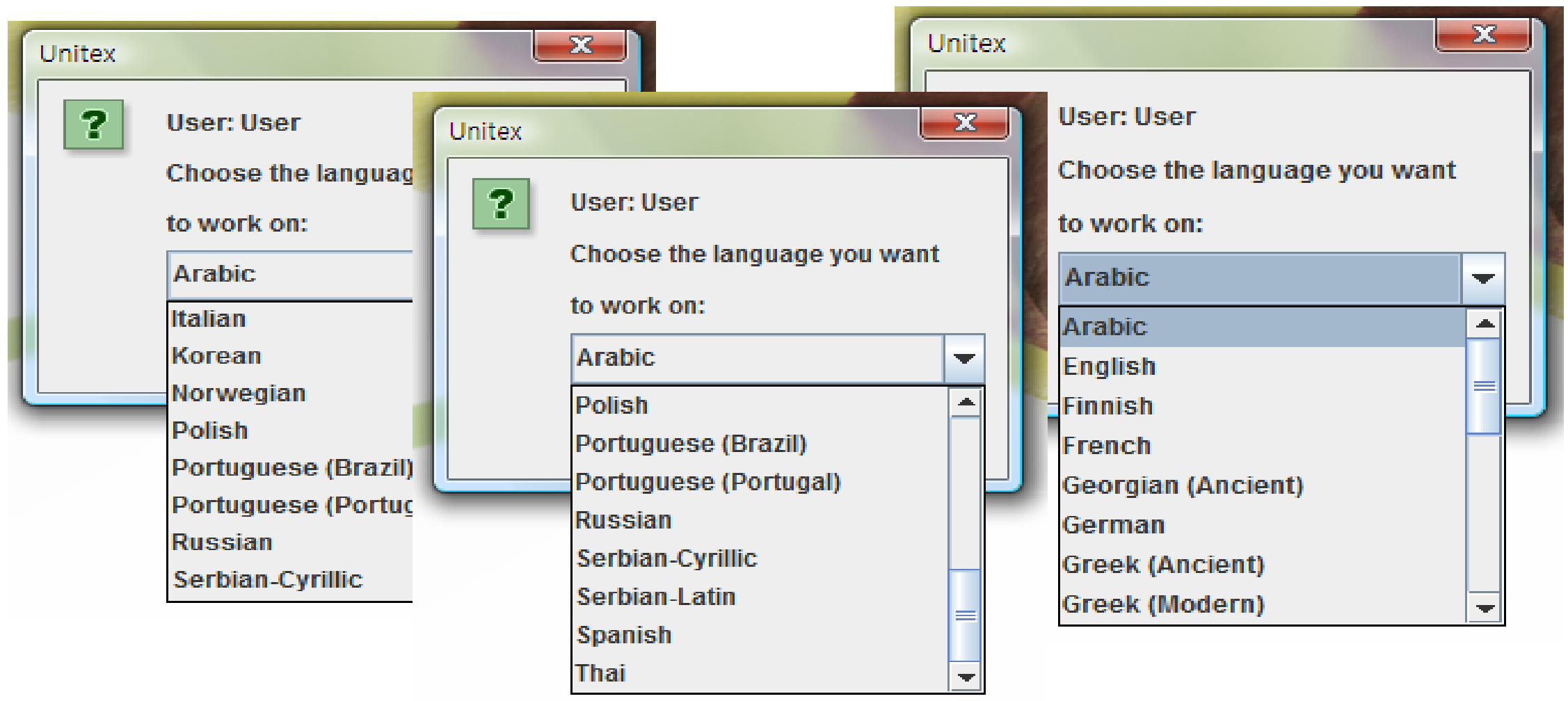

The following example represents the result of the application of an electronic dictionary to Thai text, in the form of a finite-state automaton:

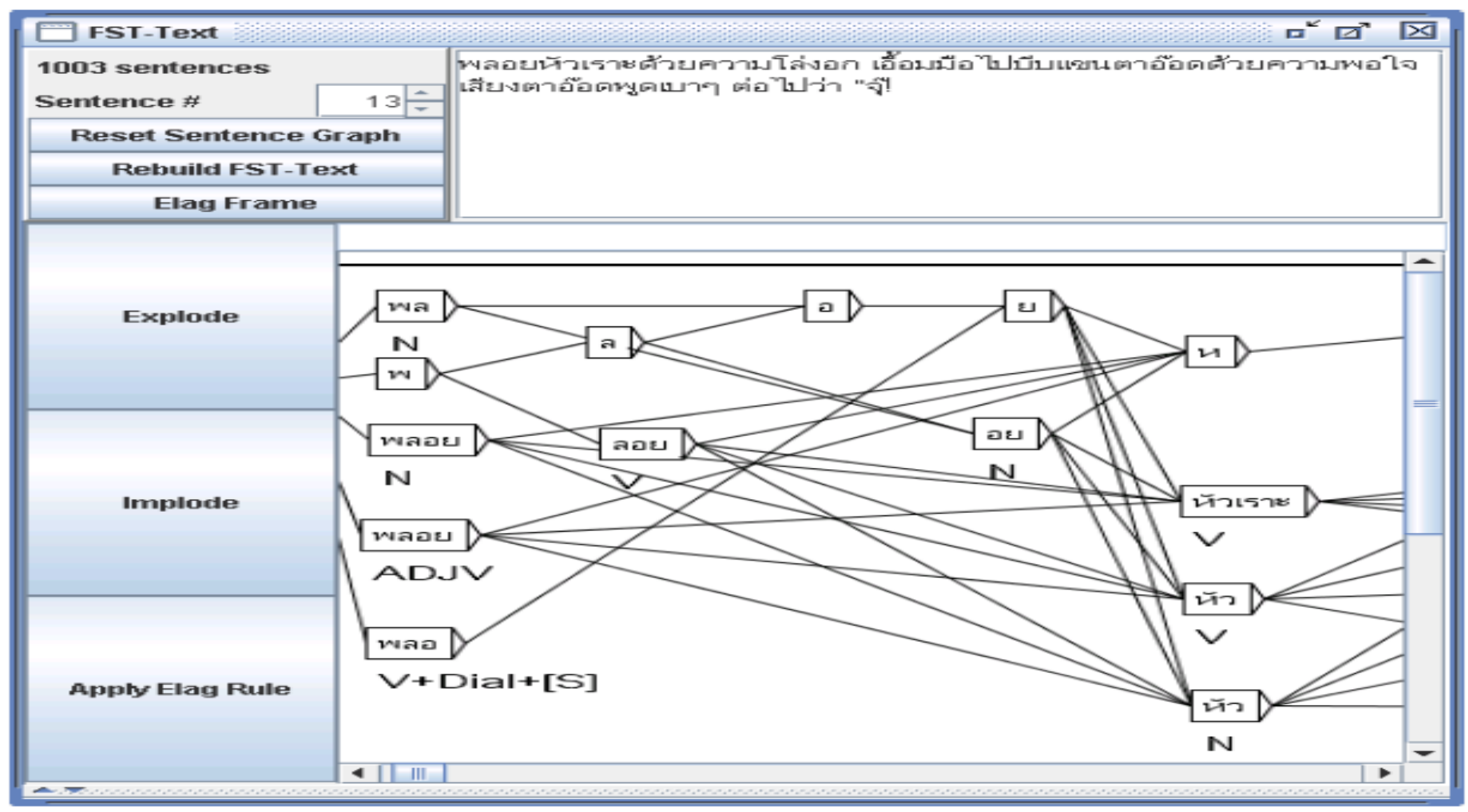

Let us now consider a Korean example. The following graph is constructed with several input words like '초콜렛[chokolles], 쪼콜레트[jjokolleteu], 초코레트[chokoleteu]', etc. and the output word '초콜릿[chokollis]' (=chocolate), the latter being the correct form.

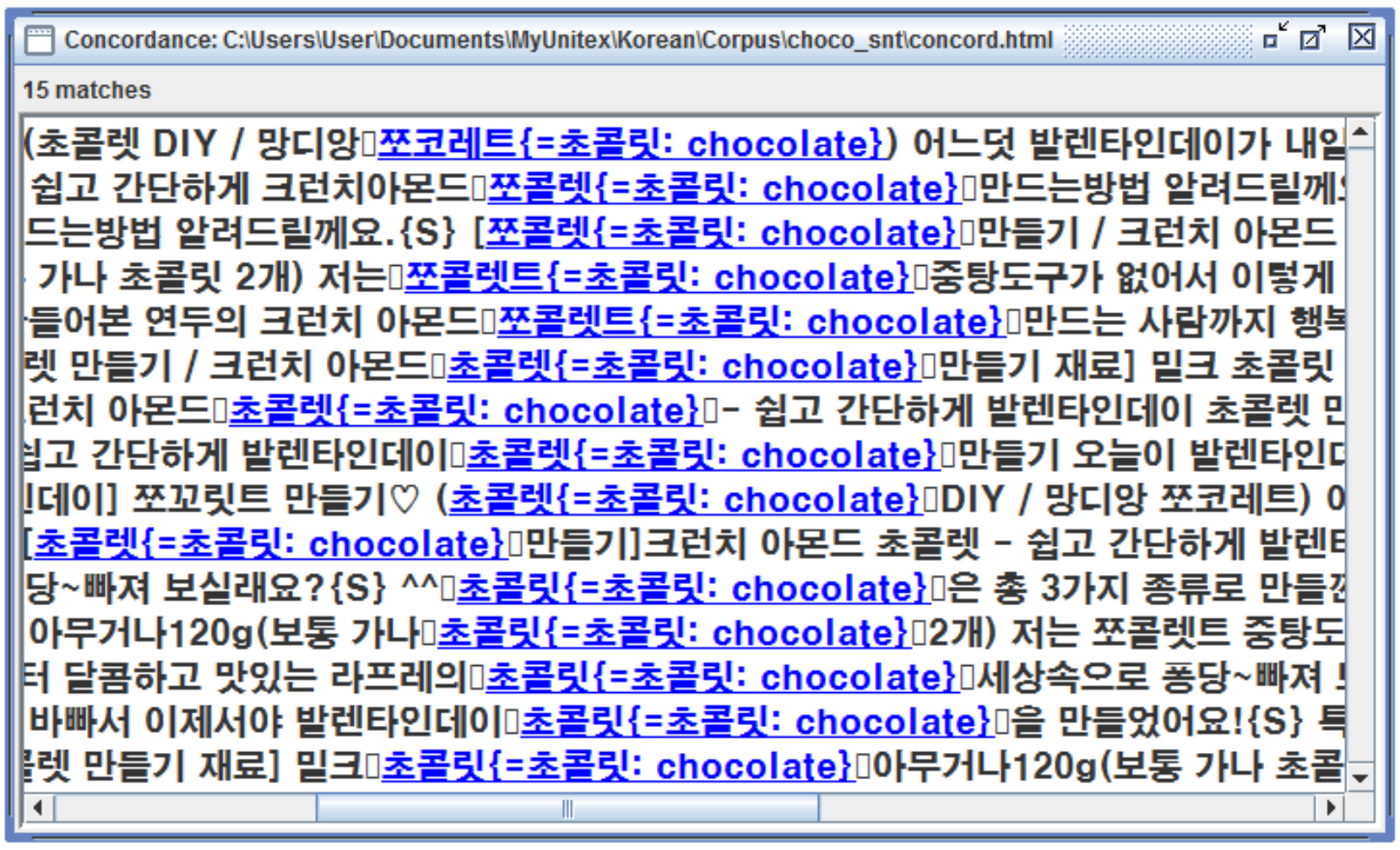

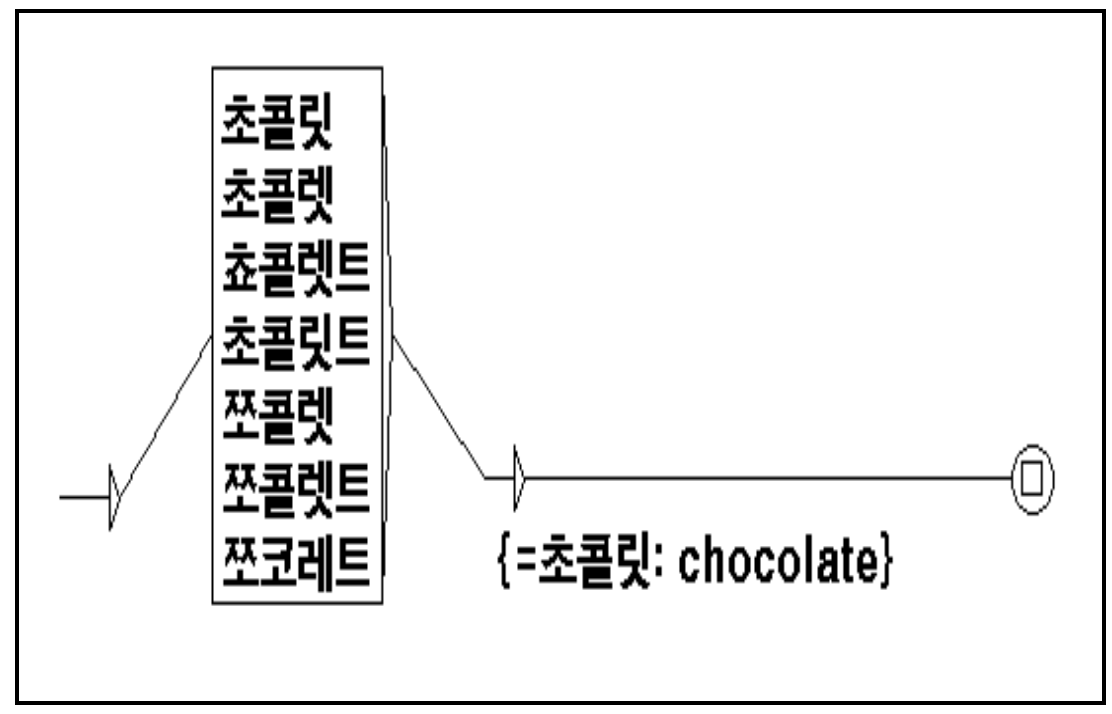

As there are many Korean transliterations of English words, LGGs like this one will help foreign learners to figure out more easily the meaning and correct spelling of a given expression.

## 5. Conclusion

To implement an effective Korean e-learning system, we built a corpus of informal web documents based on customer review texts and classified the non-standard words which are observed in these textsand commonly used in online texts in general. These non-standard words are divided into several types: expressions with missing spaces between two words, with intentional violations of orthography for saving effort or for fun, with alternative transliterations of English loan words, with newly-coined words, and finally with emoticons.

We constructed LGGs which can be integrated in a more sophisticated Korean e-learning system. Such a system would help those who learn Korean as a second language to identify and memorize these expressions with ease.